\documentclass{esann/esannV2}
\usepackage{graphicx}
\usepackage[utf8]{inputenc}
\usepackage{amssymb,amsmath,array}
\usepackage{capt-of}

\emergencystretch=\maxdimen
\begin{document}
\title{
Hierarchical Residuals Exploit \\ Brain-Inspired Compositionality
}

\author{
Francisco M. L\'{o}pez$^{1,2}$
and Jochen Triesch$^1$
%
\thanks{This research was supported by ``The Adaptive Mind'' funded by the Hessian Ministry of Higher Education, Research, Science and the Arts, Germany, and by the German Research Foundation (DFG) Project numbers 520617944, 520223571 (``Sensing LOOPS''). JT was supported by the Johanna Quandt foundation.
}
%
\vspace{.3cm}\\
%
1- Frankfurt Institute for Advanced Studies \\
Ruth-Moufang-Str. 1, 60438 Frankfurt, Germany
%
\vspace{.1cm}\\
2- Xidian-FIAS International Joint Research Center \\
Ruth-Moufang-Str. 1, 60438 Frankfurt, Germany
\vspace{.1cm}\\
\texttt{\{lopez,triesch\}@fias.uni-frankfurt.de}
}

\maketitle

\begin{abstract}
We present Hierarchical Residual Networks (HiResNets), deep convolutional neural networks with long-range residual connections between layers at different hierarchical levels. HiResNets draw inspiration on the organization of the mammalian brain by replicating the direct connections from subcortical areas to the entire cortical hierarchy. We show that the inclusion of hierarchical residuals in several architectures, including ResNets, results in a boost in accuracy and faster learning. A detailed analysis of our models reveals that they perform hierarchical compositionality by learning feature maps relative to the compressed representations provided by the skip connections. 
\end{abstract}

\section{Introduction}

The fields of artificial intelligence and neuroscience have been closely coupled since the conception of artificial neural networks (ANNs). One example of this coupling concerns the development of residual connections, which can be traced back to the pioneering work of McCulloch and Pitts \cite{mcculloch1943logical}, based on the early understanding of synaptic connectivity in biological neural networks. Residual connections, and in particular Residual Networks (ResNets) \cite{he2016deep}, are nowadays the staple for very deep ANNs because they can prevent the vanishing gradient and degradation problems by adding the activations from skip connections. ResNet-18 is commonly presented as a reasonable approximation of the visual cortex \cite{lindsay2021convolutional}. ResNeXt \cite{xie2017aggregated} is a popular variant which aggregates parallel processing paths. Residuals are also integral components of other state-of-the-art architectures such as EfficientNets and Vision Transformers; alternatively skip connections can concatenate rather than add hidden layer activations, as done in DenseNets and U-Nets.

Crucially, residual connections have also been found in the brains of insects \cite{lin2024network,winding2023connectome}, rodents \cite{siegle2021survey}, and primates \cite{yan2022mapping} -- not only skipping single layers as in ResNets but also with long-range shortcuts from early processing subcortical areas to the entire cortical hierarchy. The role of these direct connections is not fully understood but it is hypothesized that they provide fast information transfer and enable compositionality \cite{suzuki2023deep,lopez2024toward}.

In this work, we extend ANNs on the basis of these neuroanatomical discoveries about long-range residual connections. Doing so requires dealing with connections at different levels of hierarchical abstraction, i.e. compression. We propose the use of hierarchical residuals: rather than learning unreferenced representations, the deeper layers of the network need to learn feature maps relative to a straightforward compression provided by simple convolution and pooling operations. Hierarchical Residual Networks (HiResNets) introduce a marginal number of additional parameters but result in higher accuracies and faster training times than ResNets and other popular models.

\begin{figure}[t!]
\centering
\includegraphics[width=\textwidth]{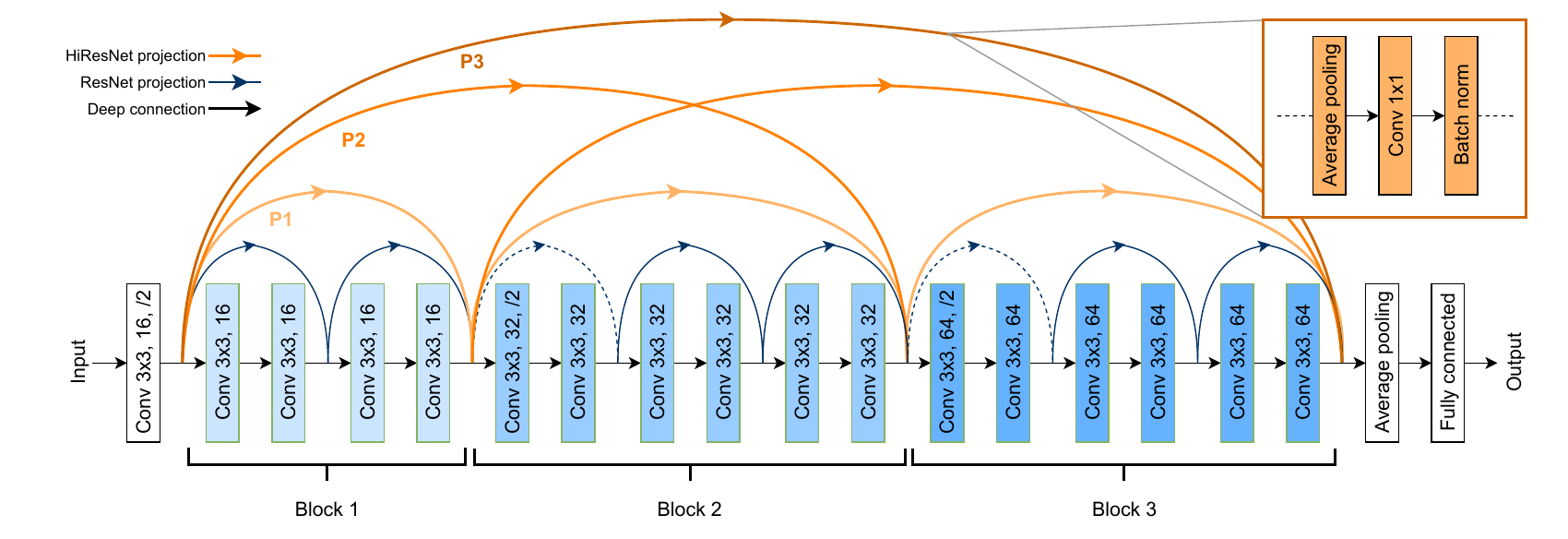}
\caption{HiResNet architecture based on the ResNet-18 with 3 residual blocks. The hierarchical residual projections consist of an average pooling to match the height and width, a \(1\times 1\) convolution to match the number of channels, and batch normalization.}\label{fig:hiresnet}
\end{figure}

\section{Hierarchical Residual Networks}

Here we provide details about the formalism of HiResNets and the innovations introduced relative to other skip connections. The output \(G(x)\) of a residual connection is the sum of a weighted connection \(F(x)\) with a projection \(P(x)\):
\begin{equation}
    G(x) = F(x) + P(x).
\end{equation}

If \(P\) is an identity mapping skip connection, i.e. \(P(x)=x\), then only the residual function \(F(x) = G(x) - x\) relative to the input needs to be learned. This is the case for the basic block in ResNets, where \(F\) is typically a stacking of two convolutional layers. The other common projection function is a \(1\times1\) convolution, which can account for mismatches in the number of channels between the input and output of a residual block.

In HiResNets, the same principles are followed to include long-range connections. First, a network is divided into blocks, within which the height, width, and number of channels are commonly preserved (see Fig.~\ref{fig:hiresnet}). We use hierarchical residual connections between blocks, such that the output of the \(\ell\)-th block becomes:
\begin{equation} \label{eq:compresnet}
    G(x_l | x_{\ell-1}, x_{\ell-2},...,x_0) = F(x_\ell) +  P_1(x_{\ell-1}) + P_2(x_{\ell-2}) + ... + P_{\ell}(x_0) \,,
\end{equation}
\noindent where \(P_k(x_{\ell-k})\) is a skip connection from the \((\ell-k)\)-th to the \(\ell\)-th block. All projection connections include \(1\times1\) convolutions and batch normalization, potentially preceded by an average pooling layer if the spatial dimensions of the inputs differ. This allows the residual function \(F\) to learn referenced representations relative to the straightforward compressions of the hierarchical projections. Because of the pooling and \(1\times1\) convolutions, each of these projections introduces only \(c_{\ell-k} \times c_{\ell}\) new parameters, where \(c_\ell\) is the number of channels of the block.

\section{Experiments}

\begin{table}[t!]
  \centering
  \begin{tabular}{lccccc}
    \hline
     & \textbf{\#P} & \textbf{CIFAR-10} & \textbf{CIFAR-100} & \multicolumn{2}{c}{\textbf{Tiny-ImageNet}} \\
    \textbf{Version} & \(\times 10^3\) & Top-1 & Top-1 & Top-1 & Top-5 \\ 
    \hline
    \multicolumn{6}{c}{Baseline architecture: Plain} \\
    \hline
    Original & \(265.0\) & \(86.50 (0.33)\) & \(55.55 (0.44)\) & \(36.53 (1.16)\) & \(62.98 (0.80)\)\\
    Adjusted & \(270.7\) &\(87.00 (0.49)\) & \(55.25 (1.00)\) & \(36.95 (1.23)\) & \(63.85 (0.91)\)\\
    HiResNet & \(270.7\) &\(\mathbf{88.27 (0.23)}\)  & \(\mathbf{58.74 (0.88)} \)& \(\mathbf{42.22 (0.37)}\) & \(\mathbf{69.08 (0.38)}\)\\
    \hline
    \multicolumn{6}{c}{Baseline architecture: ResNet} \\
    \hline
    Original & \(267.8\) & \(89.15 (0.31)\) & \(60.83 (0.56)\) & \(43.31 (0.83)\) & \(69.46 (0.67)\)\\
    Adjusted & \(273.5\) & \(89.16 (0.19)\) & \(61.25 (0.23)\) & \(43.452 (0.73)\) & \(69.75 (0.78)\)\\
    HiResNet & \(273.5\) & \(\mathbf{89.46 (0.30)}\) & \(\mathbf{62.22 (0.40)}\) & \(\mathbf{44.00 (0.37)}\) & \(\mathbf{70.82 (0.47)}\)\\
    \hline
    \multicolumn{6}{c}{Baseline architecture: ResNeXt} \\
    \hline
    Original & 70.8 & \(85.91 (0.36)\) & \(55.07 (0.28)\) & \\
    Adjusted & 76.5 & \(86.65 (0.26)\) & \(56.24 (0.22)\) & \(41.73 (1.12) \) & \(67.28 (0.43)\)\\
    HiResNet & 76.5 & \(\mathbf{86.97 (0.16)}\) & \(\mathbf{56.51 (0.38)}\) & \(\mathbf{42.38 (1.42)} \) & \(\mathbf{67.98 (0.70)}\)\\
    \hline
  \end{tabular}
  \caption{Accuracy comparisons on different datasets. The values reported are the means and standard deviations over 5 random initializations.}\label{tab:results}
\end{table}

\subsection{HiResNets outperform baseline models}

To illustrate the benefits of using hierarchical residuals, we compare the performance of HiResNets with different baseline architectures: a plain deep network, a ResNet, and a ResNeXt with cardinality 4. All architectures have 18 layers of depth, based on the ResNet-18 \cite{he2016deep}. Since HiResNets introduce additional trainable parameters, we also train adjusted versions of the baseline models by increasing the number of channels in the hidden layers to match the total number of parameters of the HiResNet.

Each model is trained on classification tasks with CIFAR-10 and CIFAR-100 \cite{krizhevsky2009learning} and Tiny-ImageNet \cite{russakovsky2015imagenet}. Tiny-ImageNet has pre-defined training, validation, and testing datasets. For CIFAR-10 and CIFAR-100, the default testing dataset is used and 10\% of the training data is separated for validation. All models are trained on CIFAR-10 for 80 epochs, CIFAR-100 for 100 epochs, and TinyImageNet for 120 epochs. We use the Adam optimizer with a learning rate of 0.01, halved every 20 epochs, and batch sizes of 100 images.

The main results are presented in Table~\ref{tab:results}. We find that the HiResNet versions of all architectures outperform the baselines for all datasets. This advantage is most prominent when compared with the plain deep network, likely because the hierarchical residuals allow for the gradients to reach each layer of the network faster. The gains with respect to the ResNet and ResNeXt are moderate, since both models use short-range residuals. As suggested for ResNets \cite{he2016deep}, residual connections are most beneficial when used in very deep networks; the ResNet-18 baseline architecture is barely deep enough. Nevertheless, since the performance of the HiResNets is higher than the adjusted versions of the baseline architectures, this suggests that hierarchical skip connections are a more efficient way of increasing the number of parameters of a model than merely increasing the depth of the hidden layers.

\subsection{Feature maps reveal hierarchical compositionality}

Next, we investigate the origins of the performance of HiResNets. To do so, we explore the contributions of the residual and projection functions in the outputs of the three blocks. Fig.~\ref{fig:activations} shows the average absolute activations of all feature maps of each connection type for all images in the TinyImageNet test dataset. In all cases, the activations of the projections are approximately half of those of the residual functions. These results confirm that the outputs of the HiResNet are mostly driven by deeper connections provided by the residuals, but also that the model is exploiting combinations of the representations at the different levels to make its predictions, i.e. it is performing hierarchical compositionality. By way of example, at the output of the third block the hierarchical connections have comparable contributions among them, indicating that the residual is learning representations not only referenced by one of these projections but to all of them. Alternatively, these results can be interpreted as showing that the compression provided by the HiResNet projections is useful to backpropagate the loss, as it provides shortcuts for each layer when computing the gradient descent, which can quickly connect them to the error at the output layer.

\subsection{Output hierarchical residuals are most beneficial}

Finally, we evaluate the relative importance of the different hierarchical residuals by performing an ablation study. Eq. (\ref{eq:compresnet}) represents the general case in which all possible projections are used, which we refer to as the default HiResNet. However, it is unlikely that all hierarchical residuals are equally important. We hypothesize that two variants of this architecture may provide interesting computational advantages inspired by the subcortico-cortical shortcut connections of the brain: the HiResNet-In introduces residuals from the first convolutional layer to all residual blocks; the HiResNet-Out introduces residuals from all residual blocks to the last one before the output. 

\begin{figure}[t!]
\begin{minipage}[c]{.5\textwidth}
  \centering
  \begin{tabular}{lcc}
    \hline
     & \textbf{\#P} & \textbf{CIFAR-10} \\
    \textbf{Version} & \(\times 10^3\) & Top-1 \\
    \hline
    ResNet (orig.) & \(267.8\) & \(89.15 (0.31)\) \\
    ResNet (adj.) & \(273.5\) & \(89.16 (0.19)\) \\
    \hline
    HiResNet &  \(273.5\) &\(\mathbf{89.46 (0.30)}\)\\
    HiResNet-In & \(269.6\) & \(88.96 (0.13)\)\\
    HiResNet-Out & \(272.3\) &\(89.37 (0.18)\)\\
    HiResNet-\(P_1\) & \(270.6\) & \(89.20 (0.17)\)\\
    HiResNet-\(P_2\) & \(269.6\) & \(88.78 (0.24)\)\\
    HiResNet-\(P_3\) & \(269.0\) & \(88.93 (0.23)\)\\
    \hline
  \end{tabular}
  \captionof{table}{Accuracy comparisons on ResNets and ablated versions of HiResNets.}\label{tab:ablation}
\end{minipage}%
\hfill
\begin{minipage}[c]{.4\textwidth}
  \centering
  \includegraphics[width=\textwidth]{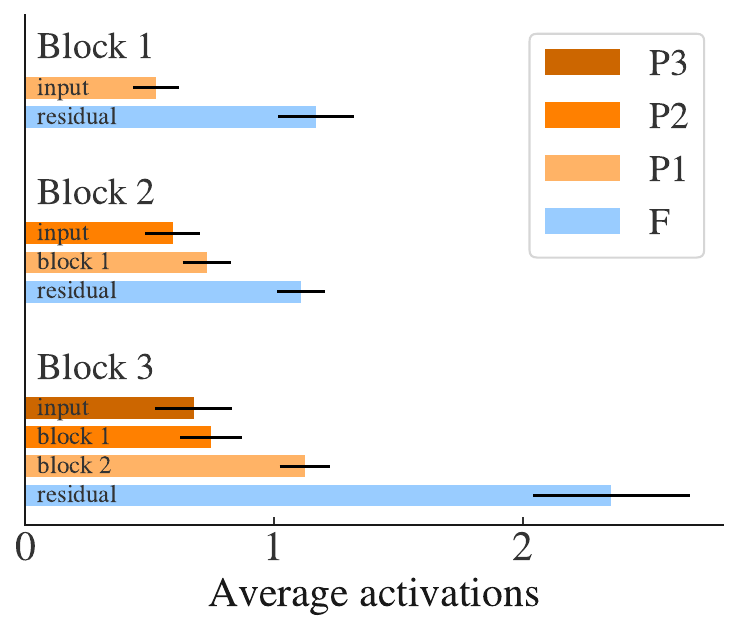}
  \caption{Average absolute activations of projections and residuals in each block.}\label{fig:activations}
\end{minipage}
\end{figure}

We repeat the CIFAR-10 experiments using the ResNet as a baseline architecture with six alternative versions of our model: the default HiResNet, the HiResNet-In, the HiResNet-Out, and additionally HiResNet-\(P1\), HiResNet-\(P2\), and HiResNet-\(P3\), which only include the first, second, and third hierarchical residuals, respectively. As shown in Table~\ref{tab:ablation}, the full HiResNet achieves the best performance. However, among the ablated versions, the HiResNet-Out is nearly as accurate as the full model, despite having fewer parameters. On the other hand, the \(P1\) version achieves comparable accuracy to the ResNets, whereas the HiResNet-In and the \(P2\) and \(P3\) versions all produce lower performances. Therefore, we conclude that while all skip connections may be of use for the HiResNet, the residuals that connect the input to the output faster are most beneficial, likely because they enable better hierarchical compositionality. This will be of particular relevance if hierarchical residuals are used in larger models, where adding too many skip connections can result in an excessive increase in the amount of parameters.

\section{Conclusion}

In this work, we introduce the Hierarchical Residual Network (HiResNet), a brain-inspired architecture with long-range residual connections between blocks at different hierarchical levels. HiResNets can outperform ResNets and other ANNs, achieving higher classification accuracies and faster learning times, by exploiting hierarchical compositionality. These results reveal that long-range shortcut connections are a more efficient method to expand the number of parameters of a model than to increase the layer depth. Further experiments are required to determine whether hierarchical residuals are also beneficial in very deep networks, e.g. ResNet-101, and other classes of architectures, e.g. Vision Transformers, potentially by increasing the complexity of the shortcut connections. Furthermore, it remains to be seen whether the biological inspiration of the HiResNet translates into this model better reproducing hierarchical processing effects of biological neural networks, e.g. incongruences between low- and high-level features. This work shows that the coupling between artificial intelligence and neuroscience is still worth exploring, as there is much potential to be gained by drawing inspiration from biological brains.


\begin{footnotesize}
\bibliographystyle{unsrt}
\bibliography{references}
\end{footnotesize}


\end{document}